\documentclass[journal]{IEEEtran}
\ifCLASSINFOpdf
\else
\fi

\usepackage{times}
\usepackage{epsfig}
\usepackage{graphicx}

\usepackage{amssymb}

\usepackage{mathtools}
\usepackage{color}

\usepackage{cases}
\usepackage{booktabs}
\usepackage{multirow}
\usepackage{multicol}
\usepackage{algpseudocode}
\usepackage{graphicx}
\usepackage{makecell}

\usepackage{arydshln}
\usepackage{amsmath}
\usepackage{algorithm}
\begin{document}

\title{Real-time Memory Efficient Large-pose Face Alignment via Deep Evolutionary Network 
}


\author{Bin~Sun,~
        Ming~Shao,~ \IEEEmembership{Member,~IEEE,}
        Siyu~Xia,~ \IEEEmembership{Member,~IEEE,}
        and~Yun~Fu,~\IEEEmembership{Fellow,~IEEE}
\thanks{Bin Sun and Yun~Fu are with the Department of Electrical and Computer Engineering, Northeastern University, Boston, MA, USA. Ming Shao is with the Institute of CIS, College of Engineering, University of Massachusetts Dartmouth, North Dartmouth, MA, USA. Siyu Xia is with School of Automation, Southeast University, Nanjing, China. (e-mail:sun.bi@husky.neu.edu; mshao@umassd.edu; xia081@gmail.com; yunfu@ece.neu.edu )}
}

\markboth{Journal of \LaTeX\ Class Files,~Vol.~14, No.~8, October~2019}%
{Shell \MakeLowercase{\textit{et al.}}: Bare Demo of IEEEtran.cls for IEEE Journals}
\maketitle

\begin{abstract}
There is an urgent need to apply face alignment in a memory-efficient and real-time manner due to the recent explosion of face recognition applications. However, impact factors such as large pose variation and computational inefficiency, still hinder its broad implementation. To this end, we propose a computationally efficient deep evolutionary model integrated with 3D Diffusion Heap Maps (DHM). First, we introduce a sparse 3D DHM to assist the initial modeling process under extreme pose conditions. Afterward, a simple and effective CNN feature is extracted and fed to Recurrent Neural Network (RNN) for evolutionary learning. To accelerate the model, we propose an efficient network structure to accelerate the evolutionary learning process through a factorization strategy. Besides, we also propose a fast recurrent module to replace the traditional RNN for real-time regression. Extensive experiments on three popular alignment databases demonstrate the advantage of the proposed models over the state-of-the-art, especially under large-pose conditions. Notably, the computational speed of our model is $6$ times faster than the state-of-the-art on CPU and 14 times on GPU. We also discuss and analyze the limitations of our models and future research work.
\end{abstract}

\begin{figure*}[t]
\begin{center}
\includegraphics[width=1\linewidth]{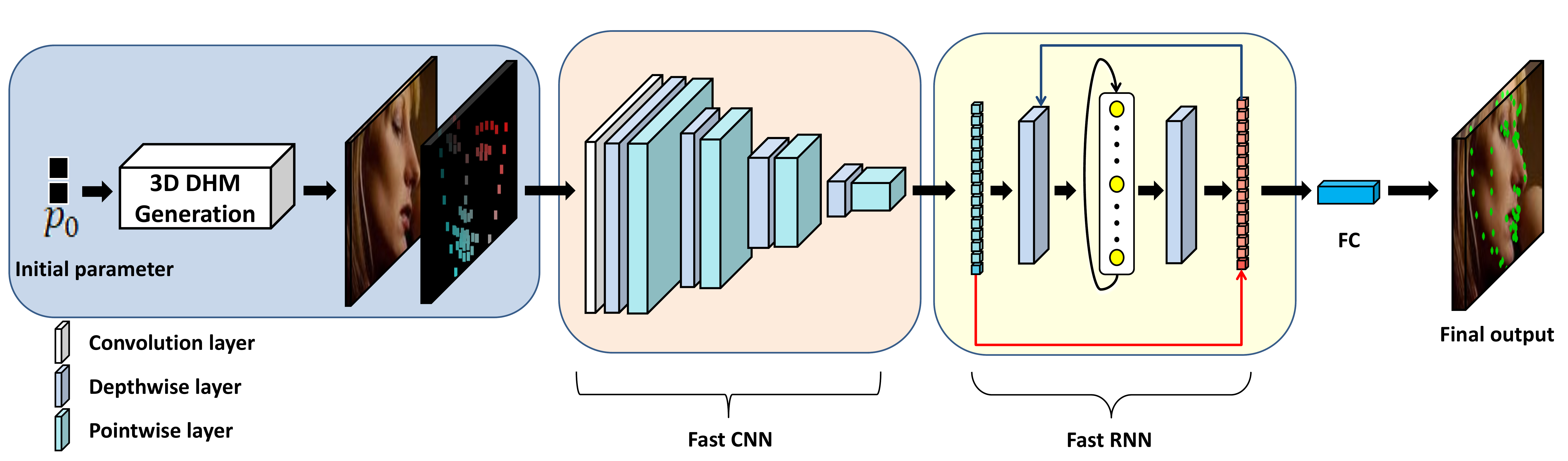}
\end{center}
\caption{Framework overview. Different from our previous work \cite{Sun2018DeepE3}, 
this framework only uses RNN as its evolutionary regression, and its output is the normalized coordinates of landmarks. Through this improvement, the new framework in the paper could significantly accelerate our previous work \cite{Sun2018DeepE3} while reducing model size. A \textbf{fast CNN} module constructed by pointwise convolution and depthwise convolution is utilized in the framework. We also propose a \textbf{fast RNN} structure to further reduce the time complexity and space complexity of the whole framework. }
\label{fig:struct}
\end{figure*}

\section{Introduction}
Face recognition and related application become increasingly popular, especially with the advances of deep learning. To name a few, face identification/verification \cite{zhao2003face}, gaze detection \cite{hansen2010eye}, virtual face make-up \cite{guo2009digital}, age synthesis \cite{fu2010age}, etc. Almost all of them heavily rely on face alignment that automatically locates predefined key points on a face. It has been treated as one of the fundamental problems in real-world face recognition systems. Nonetheless, there are still many barriers to face alignment techniques, involving unconstrained facial poses, complex expressions, and variable lighting conditions \cite{bazrafkan2018face}. Among these factors, large facial pose is of great impact \cite{3DSTN} and possible reasons are: (1) Face alignment relies on discriminant features, and these features become unreliable when they are from hidden landmarks \cite{Sun2018DeepE3}; (2) Conventional 2D models become vulnerable in modeling invisible landmarks due to its less flexibility; (3) Labeled 2D data is not reliable because invisible landmarks may be roughly estimated only according to the principle of symmetry.  

In recent years, deep learning networks have shown great ability to overcome this problem by utilizing 3D information \cite{liuxiaoming2016joint,bulat2017far,Sun2018DeepE3}. Our previous work \cite{Sun2018DeepE3} also shows promising performance on faces of large pose variations. However, the time complexity and space complexity of deep convolution methods often beyond the capabilities of many mobile and embedded devices \cite{mobilenets,Mobilenetv2,zhang2017shufflenet}. Therefore, reducing the computational cost and storage size of deep networks becomes an essential and challenging task for further application.

In this work, we mainly focus on two challenges in recent face alignment research: \textit{invisible features} and \textit{high computational cost}. To summarize, the contributions of this paper are:
\begin{itemize}

\item We propose a simple yet robust alignment feature learning paradigm using 3D Diffusion Heap Maps (DHM) and CNN to create high-level, reliable features containing both 2D and 3D information. Note that our DHM is calculated from the 3D model and only has three channels while 3DFAN \cite{bulat2017far} has 68 channels. This reduces the computation cost significantly.

\item We cast face alignment to a deep evolutionary model with both 2D texture and 3D structure. Specifically, we use RNN to model the dynamics of the least square system. The system overview can be found in Figure \ref{fig:struct}. 

\item  To achieve real-time performance, the output of each iteration in RNN module \cite{Sun2018DeepE3} is changed from parameters for 3D faces to normalized coordinates of landmarks. We further investigate factorized convolution structure \cite{mobilenets} to accelerate our feature extraction process and keep the performance from a recession. 

\item  We propose a fast recurrent module for our evolutionary learning paradigm  as the light-weight replacement of traditional RNN module. Comparing with previous work \cite{Sun2018DeepE3}, the fast recurrent module achieves even better result using much less computation cost and parameters.

\item We conduct extensive experiments and improve the performance on a few benchmarks. We outperform the state-of-the-art by a large margin and show the robustness on the original AFLW2000-3D dataset \cite{zhu2016face} and LS3D-W \cite{bulat2017far}. Besides, a detailed comparison of parameters and speed is demonstrated in this work to show the efficiency of our model.

\item Although this paper is an extension of our previous work published in \cite{Sun2018DeepE3}, there exist some significant differences between the two works. Compared with the previous one, this framework is more efficient by utilizing the depthwise separable CNN structure and a fast RNN structure which are firstly proposed in this paper. To further reduce the computation cost, the output of this framework is landmarks whereas the output of previous work \cite{Sun2018DeepE3} is the parameters for the BFM model \cite{blanz2003face}.
\end{itemize}

\section{Related work}

\textbf{2D Face Alignment:} The first milestone work of 2D face alignment is ASM \cite{cootes1995active}, followed by many successful non-deep algorithms \cite{cootes2001active,baltruvsaitis20123d} that considered the local patches around the facial landmarks as the features. Recently, critical works include tree-based models \cite{kazemi2014one,ren2014face} which improved the speed of face alignment to more than 1000 frames per second. Xiong et al. demonstrated the Supervised Descent Method (SDM) \cite{xiong2013supervised} with the cascade of weak regressors for face alignment, and achieved the state-of-the-art performance. Zhu et al. extended the work \cite{zhang2014coarse} and presented a new strategy \cite{zhu2015face} for large poses alignment by searching the best initial shape. Sun et al. \cite{sun2013deep} firstly employed CNN model for face alignment tasks with a raw face as the input and conduct regression with high-level features. Another extension of SDM called Global Supervised Descent Methods (GSDM) \cite{xiong2015global} tried to solve the large poses problem by dividing the training space into different descent spaces. All these face alignment methods only use 2D information and most of them use cascade method \cite{kazemi2014one,xiong2013supervised,zhang2014coarse} and local patch features \cite{xiong2013supervised,baltruvsaitis20123d,cootes2001active,trigeorgis2016mnemonic}. SAN \cite{SAN2018style} implemented GAN model to generate training sample for further improvement of the accuracy. DSRN \cite{DSRN2018direct} utilized doubly CNN and Fourier embedding to learn the low-rank features of the facial key points. Differently, we suggest an integration of global 2D and 3D deep evolutionary network to overcome the information loss caused by 2D patch features.

\textbf{3D Face Alignment:} As 3D face model can maintain the depth information well against pose issues, a bunch of 3D face alignment methods and 3D face datasets become increasingly popular. Doll{\'a}r et al. \cite{dollar2010cascaded} estimated the landmarks on large poses through a 3D Morphable model (3DMM) with cascade regression in 2010. Zhu and Jourabloo et al. \cite{zhu2016face,jourabloo2016large} presented CNN based model fitting a 3DMM to a 2D face through a cascade method, along with the facial key points. A very dense 3D alignment model has been demonstrated by Liu et al. \cite{liu2017dense,pifa} and achieved good performance. A 3D face training dataset called 300W-LP, and a testing dataset called AFLW-20003D were offered in \cite{zhu2016face} recently. Another 3D alignment dataset called LS3D-W was published by Bulat et al. \cite{bulat2017far} with about 230,000 images, and the deep learning models based on this dataset, e.g., HourGlass(HG) \cite{bulat2017far,yang2017stacked,binarizedhg}, have achieved impressive performance very recently.  In this paper, we use sparse 3D heat maps together with the original image as input, whereas most of the previous works use dense 3D models. Jointly working with a plain RNN and a simple CNN, our model achieves the state-of-the-art performance.

\begin{figure*}[t]
\begin{center}
\includegraphics[width=1\linewidth]{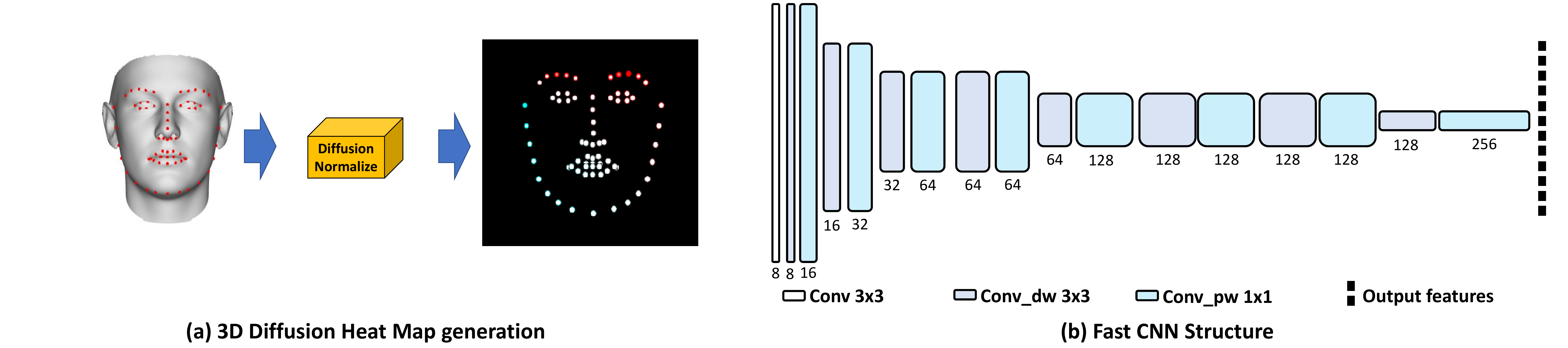}
\end{center}
\caption{Illustration of (a) generating 3D diffusion heat maps; (b) fast CNN module used as part of the new fast DHM framework.}
\label{fig:3Dheatmap}
\end{figure*}

\textbf{Efficient Network:} In recent years, more efforts have been taken to speed up the deep learning models. Faster activation function named rectified-linear activation function (ReLU) was first proposed to accelerate the model \cite{relu}. In \cite{sifre2014depthconv} depthwise separable convolution was initially introduced and was used in Inception models \cite{ioffe2015batch}, Xception network \cite{chollet2016xception}, MobileNet \cite{mobilenets,Mobilenetv2}, and Shufflenet \cite{zhang2017shufflenet}. Jin et.al. \cite{jin2014flattened} showed the flattened CNN structure was able to accelerate the feedforward procedure. A Factorized Network \cite{DBLP:factorized} was designed with a similar philosophy as well as the topological connection. A compression method of a deep neural network was introduced in \cite{ba2014deep}, indicating that in certain cases complicate deep models are equal in performance by small models. Then Hinton et al. extended the work in \cite{hinton2015distilling} with the weight transfer strategy. Squeezenet \cite{iandola2016squeezenet} combined such work with a fire module containing many $1 \times 1$ convolutional layers. Another strategy \cite{binarized,xnor} of converting the parameters from float type to binary type can compress the model significantly and achieve an impressive speed. However, the performance will be compromised. In this paper, we use the factorized convolution structure in RNN modeling to reduce the number of parameters and accelerate the model. 


\section{Algorithms}

In this section, we will detail our new framework (Figure \ref{fig:struct}) including three exclusive components: (1) 3D DHM generation; (2) deep evolutionary 3D heat maps; (3) fast deep evolutionary framework.

\subsection{3D Diffusion Heat Maps}

To get feasible 3D landmarks in large poses, one of the reasonable ways is to build a 3D model of the face and simulate the details of a real face such as scale, expressions, and rotations, which can be formulated by the state-of-the-art 3DMM \cite{blanz2003face}. Typically, it represents factors of a 3D face by:
\begin{equation}
S=\overline{S}+E_{\text{id}}p _{\text{id}}+E_{\text{exp}}p_{\text{exp}},
\label{equ.1}
\end{equation}
where $S$ is a predicted 3D face, $\overline{S}$ is the mean shape of the 3D face, $E_{\text{id}}$ is principle axes based on neutral 3D face, $p_{id}$ is shape parameters, $E_{\text{exp}}$ is principle axes based on the increment between expressional 3D face and neutral 3D face, and $p_{\text{exp}}$ is expression parameters. In our framework, the $E_{\text{id}}$ and $E_{\text{exp}}$ are calculated from a popular 3D face model named BFM \cite{blanz2003face}. Then we project the 3D face model by Weak Perspective Projection:
\begin{equation}
F(p)=f\times M\times R\times S+t _{2\text{d}},
\label{equ.2}
\end{equation}
where $F(p)$ is the projected 3D face model, $f$ is the scale, $M$ is orthographic projected matrix $\left(
\begin{array}{ccc}
1 & 0 & 0 \\
0 & 1 & 0 
\end{array}
\right)$, $R$ is the rotation matrix written in $\left[
\begin{array}{ccc}
r_{\text{pitch}} &r_{\text{yaw}} & r_{\text{roll}} 
\end{array}
\right]$ , $S$ is the 3D shape model calculated from Eq. \eqref{equ.1}, and $t _{\text{2d}}$ is the transition vector with the location coordinate $x$ and $y$. Since $F$ is the function of the parameter $p$, $p$ can be written as $p=\left[
\begin{array}{ccccc}
f &  R& t_{\text{2d}} &p _{\text{id}}  & p _{\text{exp}}
\end{array}
\right]$. We can generate the aligned 3D shape through Eq. \eqref{equ.1} and Eq. \eqref{equ.2}. Afterwards, with the key point index provided by BFM, we will have precise locations of key points on the 3D model. To generate sparse 3D features, we normalize the coordinates in the 3D model around the key points. For a specific color channel $i$, the process could be described as:
\begin{equation}
\text{map}_{i}(k)=\frac{S_{{j}}(k)}{\text{max}(S_{{j}})-\text{min}(S_{{j}})},~~ {j}\in\{x,y,z\},
\label{equ.3}
\end{equation}
where ``$\text{map}_i()$'' is the 3D heat map with three channels R, G, and B, and the 3D coordinates triplet $\{x,y,z\}$ are mapped to the three channels. $S_{{j}}(k)$ means the value of the 3D shape at the location of the $k$-th landmarks. To incorporate the locality and increase the robustness of each landmark, we suggest to extend the 3D heat maps by a Gaussian diffusion map centered at each landmark's location, and thus, we obtain 3D DHM for robust representation. Specifically, we generate a set of heat maps centered at the landmarks and ranged by a 2D Gaussian with standard deviation of 1 pixel (See Figure \ref{fig:3Dheatmap}(a)). This strategy assures our framework can extract the features of the whole image with different weights instead of discarding the features located at a non-landmarks position. Note we conduct the normalization and diffusion on each single channel/axis, independently. A mean initialization DHM is generated before the training process.

\textbf{Discussions:}  Recall the basic inputs of existing models usually include two separate parts: (1) an image; (2) initial mean landmarks. These methods usually depend on the initial landmarks for facial features for better performance, e.g., the features are usually extracted by cropping the image centered at initial landmarks. As discussed earlier, this strategy may degrade the performance in large pose situations, as the features centered at initial landmarks have significantly deviated from the ground truth. See the ``Input'' in Figure \ref{fig:struct} (with face image and heap map). Thus, these features will misguide the learning model or regressor and probably converge to local trap. In contrast, we design a novel paradigm to address this issue. Namely, we keep the whole image as the input for robust facial feature learning and propose to employ sparse 3D shape information as the compliment. This avoids the issues of misguidance by the incorrect initial landmarks that propagate to the local features. 

In the work 3DDFA \cite{zhu2016face}, a diffusion features named PNCC was proposed. PNCC also fused 3D information with 2D image as the input. Compared with the PNCC features, our 3DDHM emphasize the features around the possible location of facial key points based on the mean shape. With such attention, our framework can localize the landmarks accurately using limited computation resource.

\subsection{Deep Evolutionary Diffusion Heat Maps}

\begin{figure*}[t]
\begin{center}
\includegraphics[width=1\linewidth]{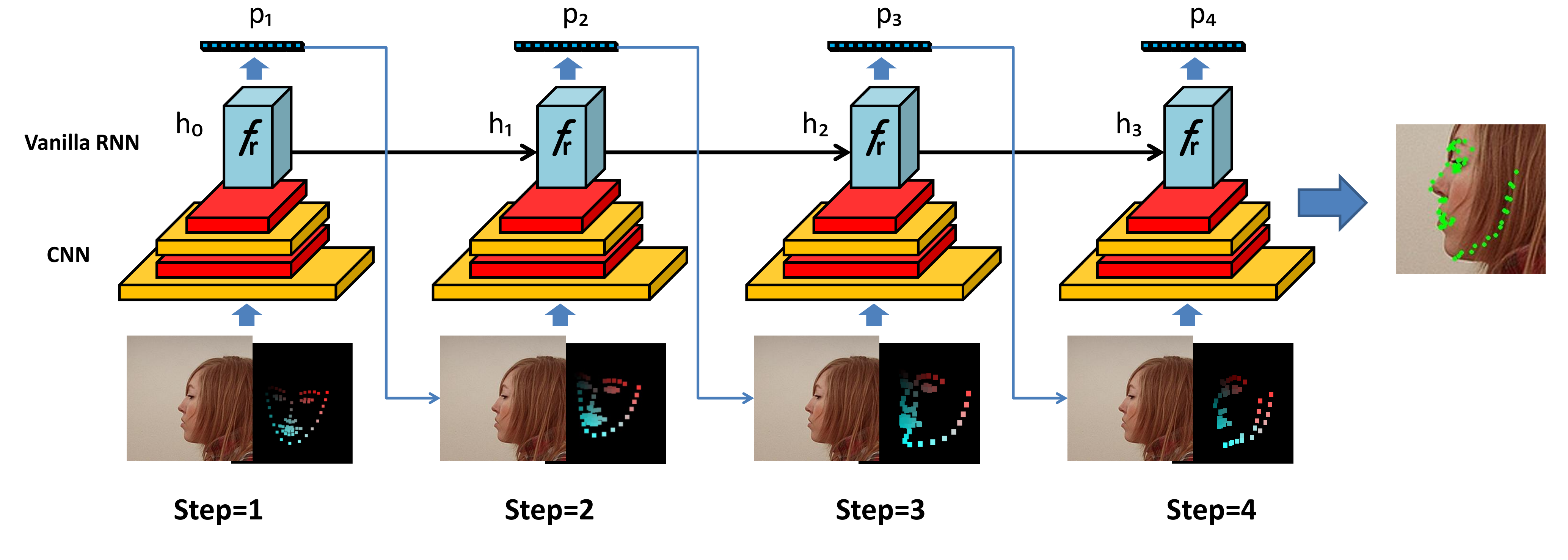}
\end{center}
\caption{Illustrations of the intermediate results of Deep Evolutionary Diffusion Heat Maps with state number $t=4$ where changes of the heat maps can be seen. The heat map is scaled heavily in step 1 and deformed in step 2 to 4 with changed color. Finally, the heat map is well aligned on the face as shown in the output.}
\label{fig:dynamicresults}
\end{figure*}

We offer an end-to-end trainable deep structure for face alignment in this section given the 3D heat maps and stacked image $I$. Since we have already generated DHM using 3D landmarks, we concentrate on: (1) discriminative and robust representation of image plus 3D DHM; (2) 3D DHM evolution; (3) recurrent HourGlass framework.

First, to formulate both discriminative and robust alignment features, we propose to use a plain CNN that absorbs both 2D and 3D information in a stacked structure. While handcraft features or direct use of 3D information may work, our practice reflects that they are less competitive than the well developed CNN model. The CNN model here can extract high-level features critical to alignment, and we are especially interested in global CNN features. In our experiments, we also find that an off-the-shelf CNN model such as VGG-net \cite{simonyan2014very} works reasonably well in our case. We use reduced VGG network to fuse the 3D information and RGB information which proves useful in our experiments.

\begin{algorithm}[t]
  \caption{Deep 3D Evolutionary Diffusion Heat Maps}
  \begin{algorithmic}
    \State \textbf{Inputs:} Image $I$, initial values: $p_{0}=[p _{\text{exp}},p _{\text{id}},f _{0},R _{0}, t _{{2d0}}]$
    \For  {$i=0$ to $\text{IterNum}$} 
    \State generate $S _{i}$ in 3D
    \State $map=\text{zeros}()$
    \For  {$j=0$ to $3$}
    \State $S _{i}[j]-=\text{min}(S _{i}[j])$
    \State $S _{i}[j]/=(\text{max}(S _{i}[j])-\text{min}(S _{i}[j]))$
    \State $map[S _{i}[0],S _{i}[1],j]=S _{i}[j]$
    \EndFor 
    \State extract features $\phi^k(cat(I,map),C _{\text{conv2}})$
    \State $\hat{p}_{k} = RNN(\phi ^k(cat(I,map)))$
    \EndFor
    \State \textbf{Outputs: } $\hat{p}_{k}$
  \end{algorithmic}\label{alg:3DDHM}
\end{algorithm}

Second, we resort to evolutionary modeling for the alignment features. RNN has been widely applied to temporal data, as it is able to account for temporal dependencies. In training, RNN maintains the topology of feedforward networks while the feedback connections enable the representation of the current state of the system which encapsulates the information from the previous inputs, which can help update the parameter $p$ in the loops within the networks. Mathematically, the update rules can be written as:
\begin{equation}
\begin{aligned}
h_{t+1} = \text{tanh}(W_{{\text{ih}}}C_{\text{conv2}}([I,\textrm{map}])+W_{\text{hh}}h_{t}),
\label{eq.4}
\end{aligned}
\end{equation}
where $h_{t}$ is the hidden state of the step $t$, $C_{\text{conv2}}([I,\text{map}])$ is the convolutional output features extracted from the input image $I$ and the heat maps ``$\text{map}$" generated by Eq. \ref{equ.2}. $W_{\text{ih}}$ is the weight from input to the hidden layer and the  $W_{\text{hh}}$ is the weight from the hidden layer to the hidden layer. With the hidden features $h$, we can model the update rule for $\hat{p}$ using $\hat{p}_{t+1}=\hat{p}_{t}+W_{\text{ho}}h_{t}$,
where $\hat{p}_{t}$ is the parameters in the step $t$, $W_{\text{ho}}$ is the weight from hidden layer to the output. Thus, from Eq. \eqref{eq.4}, 
we can prove that all the parameters in the step $t+1$ are based on the state of the step $t$. Since the whole image has been engaged as the input in each step, we may rectify the errors caused by the previous steps. Besides, we have the generated heat maps to emphasize the change in the previous step so that the whole network can converge. An illustration of evolutions of 3D diffusion heat maps in four states can be found in Figure \ref{fig:dynamicresults}.

During the training, we define our loss function as
\begin{equation}
\begin{aligned}
L=\text{min} \|S_{0}+\sum_{t=1}^{T}W_{\text{ho}}h_{t}(p)-S^*  \|_{F}^{2}
\label{eq.7}
\end{aligned}
\end{equation}
where $S^*$ is the ground truth shape, $F$ indicates the matrix Frobenius norm, $T$ is the total number of the steps. The complete algorithm is shown in Algorithm \ref{alg:3DDHM}.

The optimized $p$ in the output layer will generate 3D landmarks for the test face which can be identified in the output of Figure \ref{fig:struct}. Here we suggest using Vanilla-RNN for simplicity. The evolutionary 3D DHM and intermediate results can be found in Figure \ref{fig:dynamicresults}. Note we update the parameters of the 3D model $p$ instead of the landmarks themselves as we would encourage the model to restrict the spatial relation of each landmark. The input is a $224 \times 224 \times 3$ image stacked by heat maps. The output of the RNN module is a 234-dimensional parameter, which will be cast to a 3D face model using Eq. $(2)$. We can use specific vertices to find the landmark position. Last, to explore the generality of our framework, we upgrade the CNN by a popular stacked HourGlass module, and RNN by LSTM. In later experiments, we may compare with 3DFAN \cite{bulat2017far} that only uses the HourGlass module resulting in inferior performance. This also demonstrates our evolutionary learning strategy is more generic.

\subsection{Fast Deep Evolutionary Framework}

\begin{figure*}[t]
\begin{center}
\includegraphics[width=0.9\linewidth]{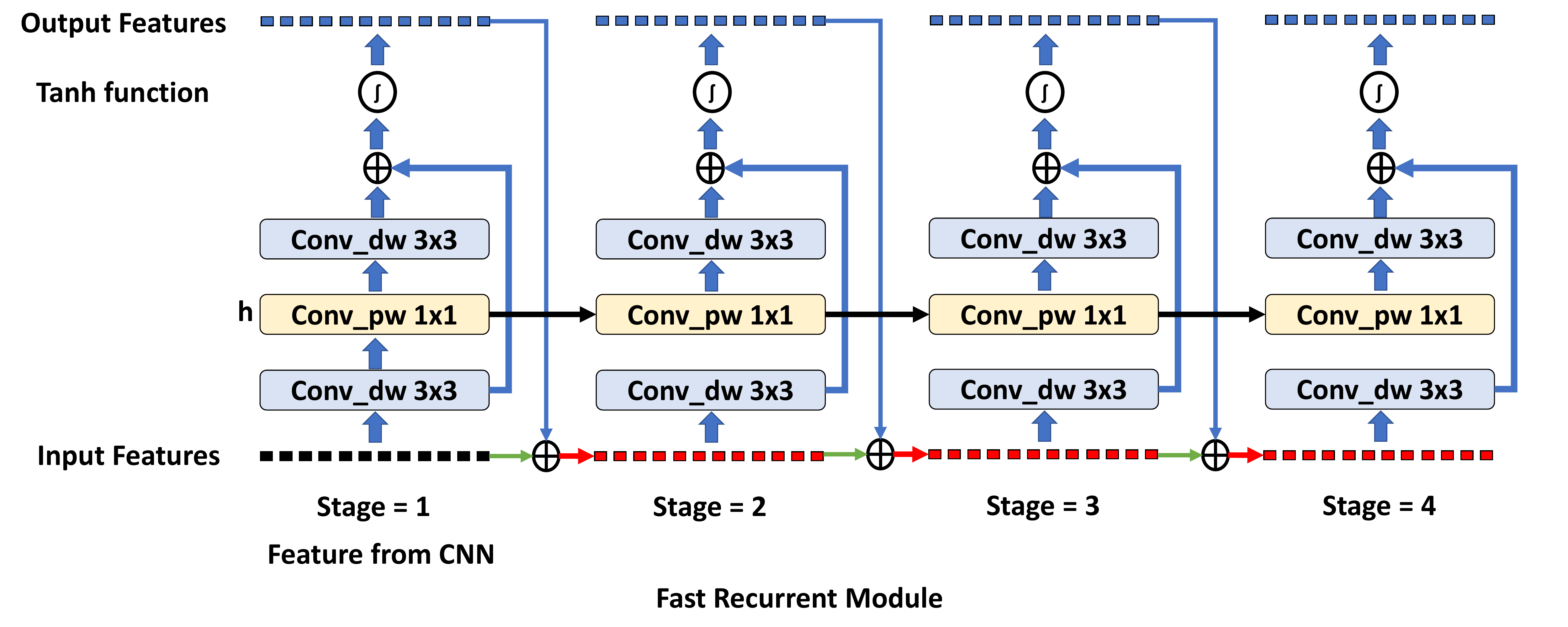}
\end{center}
\caption{The demonstration of our fast Recurrent module. Different from traditional RNN, we use one depthwise convolution layer to calculate the input weight and the output weight and use one pointwise convolution layer as the hidden layer. With our fast Recurrent module, the speed of our framework is improved from 17 FPS (with accelerated CNN module only) to 80 FPS. And as a result, the size of the model has been reduced to 976 KB.}
\label{fig:fastRNN}
\end{figure*}

An end-to-end trainable deep structure for face alignment given the 3D heat maps and stacked image $I$ is proposed before and its effectiveness has been proved \cite{Sun2018DeepE3}. However, such a framework is not efficient enough due to its iteration process. In this section, we concentrate on improving the efficiency and propose a novel framework named fast Deep Evolutionary Diffusion Heat Maps based on the Depthwise Separable Convolution module. Therefore, this subsection is divided into two parts: a brief introduction to Depthwise Separable Convolution \cite{mobilenets}, and the details of our fast Deep Evolutionary DHM framework.  
\subsubsection{Depthwise Separable Convolution}
Depthwise Separable Convolution is one of the important factorization structures which plays key roles in building many lightweight architectures \cite{zhang2017shufflenet,mobilenets,Mobilenetv2}. It consists of: (1) depthwise convolutional layer; \textbf{} (2) pointwise convolutional layer \cite{mobilenets}. 

\textbf{Depthwise convolutional layer} is to apply a single convolutional filter to each input channel. This massively reduces the computational cost and the number of parameters. Particularly, the cost can be calculated as 
$S_{\text{F}} \times S_{\text{F}} \times S_{\text{k}} \times S_{\text{k}} \times C_{\text{out}}$, where $S_{\text{F}}$ is the size of the feature map, $S_{\text{k}}$ is the size of the kernel, $C_{out}$ is the number of the output features. The amount of the parameters without bias is computed by $S_{\text{k}} \times S_{\text{k}} \times C_{out}$, where $C_{out}$ is the channel number of the output.

\textbf{Pointwise convolutional layer} is to use $1\times1$ convolution to build the new features through computing the linear combinations of all input channels. Essentially, it is a traditional convolution layer with the kernel size being $1$. Now, we show the computational cost of the traditional convolutional layer, which is calculated as $S_{\text{F}} \times S_{\text{F}} \times S_{\text{k}} \times S_{\text{k}} \times C_{\text{in}}\times C_{\text{out}}$.
Since we assume there is no bias, the parameters for the traditional layer is computed by $S_{\text{k}} \times S_{\text{k}} \times C_{\text{in}} \times C_{\text{out}}$. The fact $S_{\text{k}}=1$ in pointwise layer allows us to calculate its  computational cost as $S_{\text{F}} \times S_{\text{F}} \times C_{\text{in}} \times C_{\text{out}}$, and the parameters as $C_{\text{in}} \times C_{\text{out}}$.

Because $C_{\text{in}}$ is usually much larger than $S_{\text{k}}^{2}$ in the depthwise layer, the depthwise convolution is more efficient than pointwise convolution in terms of computational cost and parameters. 

\subsubsection{Fast DHM framework}
In brief, we accelerate both CNN and RNN models in our DHM framework. In the work of DHM \cite{Sun2018DeepE3}, a plain CNN structure is used to extract both 2D and 3D information in a stacked structure. In our fast Deep Evolutionary framework, however, such CNN structure is replaced by the depthwise separable convolution block to reduce the computation complexity and parameters. The structure is shown in Figure \ref{fig:3Dheatmap} (b). 
\begin{figure*}[t]
\begin{center}
\includegraphics[width=1\linewidth,height=0.9\linewidth]{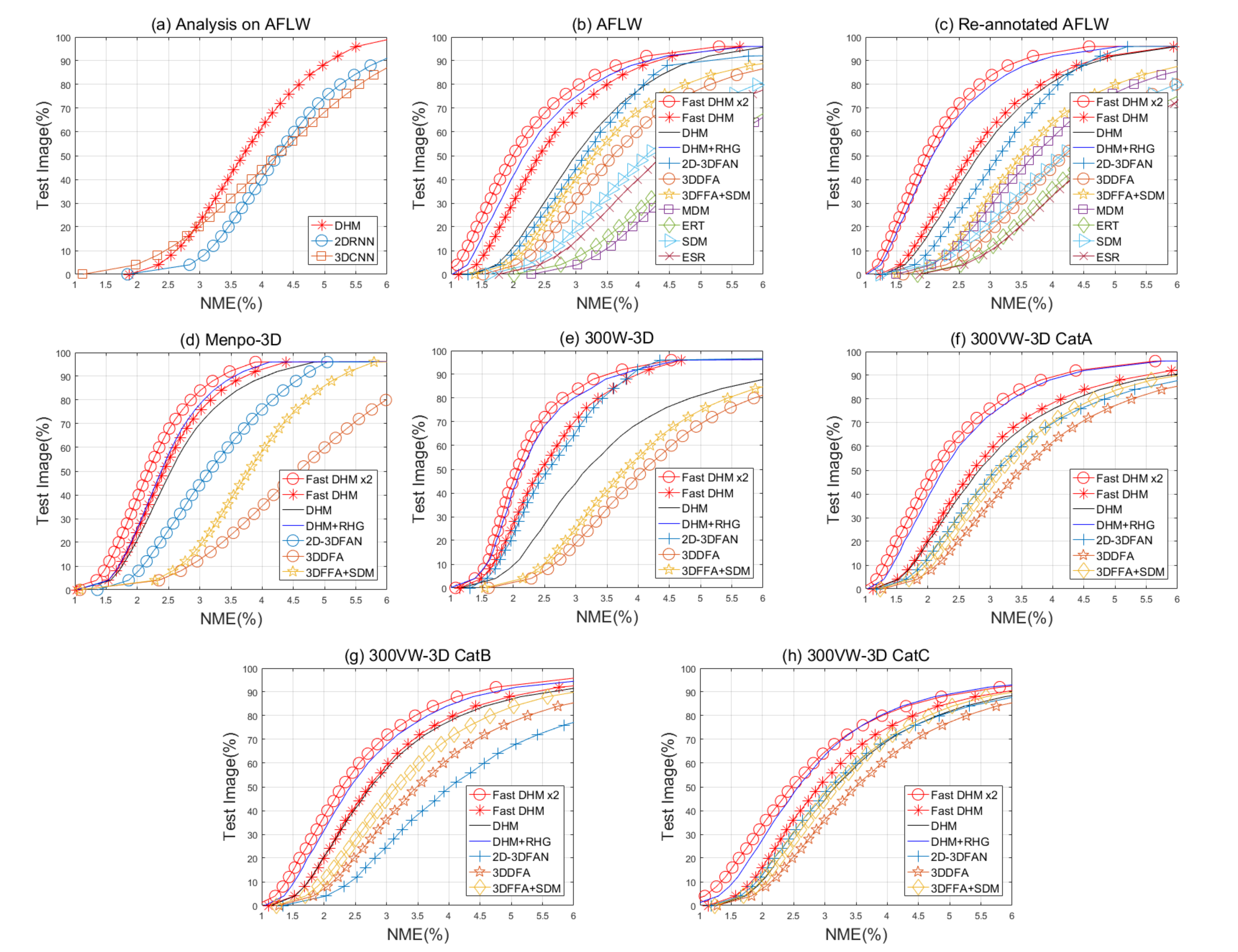}
\end{center}
\caption{
(a) shows analysis of replacing 3D diffusion heat map and RNN model. (b) to (h) show the comparisons results on different test datasets. The CED area of our fast DHM $\times 2$ is the largest among all baselines in different test datasets. The DHM with Recurrent Hourglass \cite{Sun2018DeepE3} has similar results with the Fast DHM$\times 2$, but it is much slower and larger. }
\label{fig:2DRNN&3DCNN&all}
\end{figure*}

A novel fast Recurrent Network is also proposed to further accelerate the whole framework by utilizing the depthwise convolution and pointwise convolution in the RNN structure.  The structure is shown in Figure \ref{fig:fastRNN}. Inspired by the work \cite{xiong2013supervised,trigeorgis2016mnemonic}, we use a recurrent regression to learn the optimal of the non-linear problem ${\text{min}} \|S-S^*  \|_{F}^{2}$, in which $S$ is the predicted shape and $S^*$ is the ground truth. Different from the work \cite{Sun2018DeepE3}, the output of each iteration is the feature maps for final landmarks instead of the parameter $p$ \cite{Sun2018DeepE3}. In this way, our framework can avoid the time-consuming computation of BFM in the recurrent process. However, the recurrent process is still inefficient due to the convolution blocks in its iteration procedure. To improve efficiency, the factorization convolution block is implemented in the RNN module. Mathematically, the update rules can be written as:
\begin{equation}
\begin{aligned}
F_{t+1} = F_{t}+\text{tanh}(W_{\text{ih}}F_{t}+W_{\text{hh}}h_{t}),
\label{eq.8}
\end{aligned}
\end{equation}
where $F_{t+1}$ is the output features from stage $t+1$, $F_{t}$ is the output features from stage $t$, $W_{{ih}}$ is the weight from input to the hidden layer, $h_{{t}}$ is the hidden state of the stage $t$, and the $W_{{hh}}$ is the weight from hidden layer to hidden layer. In the initial stage, the input features are the feature maps extracted from CNN structure, and the output of the pointwise convolution layer is considered as the initial hidden state $h_{0}$. The output features can be considered as the increment of the landmarks in each stage. To keep the consistency during the evolutionary process, the input features in each stage (except initial stage) is the combination of previous features and the incremental features.

To avoid the over interference of previous features, a depthwise convolution layer is implemented on the input features of the current stage. Then a depthwise separable convolution structure is adopted to replace the traditional hidden layer with the input from the previous hidden state. Following the current hidden layers, a sum function and a $tanh$ activate function is applied to generate the final incremental features. The process of Eq. \eqref{eq.8} can be written as:
\begin{equation}
\begin{aligned}
F_{{t+1}} = F_{{t}}+\text{tanh}(D_{i}\otimes F_{{t}}+D_{\text{h}}W_{\text{hh}}\otimes h_{{t}}),
\label{eq.9}
\end{aligned}
\end{equation}
where $D_{i}$ is the weight of the depthwise convolution for the input, $D_{\text{h}}$ is the weight of the depthwise convolution for the hidden state, $W_{\text{hh}}$ is the weight of the $1 \times 1$ hidden layer, and $\otimes$ is the convolution operation.



During the training, the loss function is defined as $\text{min} \|S-S^*  \|_{F}^{2}$
where $S$ is the final predicted shapes after the fully connected layer as shown in Figure \ref{fig:struct}, $S^*$ is the ground truth shape, $F$ indicates the matrix Frobenius norm.

\section{Experiments}

In this section, we will first detail the evaluation datasets for large pose face alignment and show the evaluation metrics. Then, baseline methods and parameter settings will be briefly introduced. Third, we will conduct an analysis of the proposed method and evaluate different modules. Last, we will compare with the state-of-the-art face alignment methods and analyze the results from accuracy, time complexity, and space complexity.
\subsection{Dataset and Evaluation Metrics}
We use 68-point landmarks to conduct fair comparisons with the state-of-the-art methods, though our method can adapt to any numbers of landmarks. Note in the training process, we may need 3D landmarks or parameters which are not always available. Thus, we estimate the 3D information through \cite{bulat2017far} in this situation. Evaluation datasets are detailed below:
\begin{itemize}
\item 300W-LP: The dataset has four parts, a total of 61,225 samples across large poses (1,786 from IBUG, 5,207 from AFW, 16,556 from LFPW and 37,676 from HELEN) \cite{zhu2016face}. Note we used 58,164 images for training and 3,061 as the validation.
\item AFLW2000-3D: The dataset is essentially a reconstruction by Zhu et al. \cite{zhu2016face} given 2D landmarks. Note we use it for testing with 2000 images in total.
\item Re-annotated AFLW2000-3D: The dataset is relabeled by Bulat et al. \cite{bulat2017far} from AFLW2000-3D given 2D landmarks. We use it for testing with 2000 images in total.
\item LS3D-W: The dataset is also a re-annotated by Bulat et al. \cite{bulat2017far}. We use it for training and testing to make a fair comparison. We use 218,595 images for training, and use its sub-datasets Menpo-3D, 300W-3D, 300VW-3D (A), 300VW-3D (B), and 300VW-3D (C) for testing. Note this dataset only has 2D landmarks projected from 3D space.
\end{itemize} 

As our focus is face alignment, we should reduce the negative effects of face detection. Thus, the detected bounding box of each face is computed by ground-truth landmarks. To compare with other methods, we use the same metric ``Normalized Mean Error (NME)'' defined as $\text{NME}=\frac{1}{N}\sum_{i=1}^{N}\frac{\left \|\hat{X}_{\text{i}}-X_{\text{i}}^*\right \|^{2}_{2}}{d}$ where the $\hat{X}$ and $X^*$ is predicted and ground truth landmarks, respectively, $N$ is the number of the landmarks, $d$ is normalized distance computed by the width and height of the bounding box using $d=\sqrt{w_{\text{bbox}}\times h_{\text{bbox}}}$. The lower NME means higher accuracy. We also show the curve of cumulative errors distribution (CED) and set the failure threshold as $6\%$. The CED curve indicates the percentage of successful cases in the test dataset. The speed of all methods is evaluated on Intel-i7 CPU with one core only. The storage size in this paper is calculated from the compressed model generated from the source code.

\begin{figure*}[t]
\begin{center}
\includegraphics[width=1\linewidth]{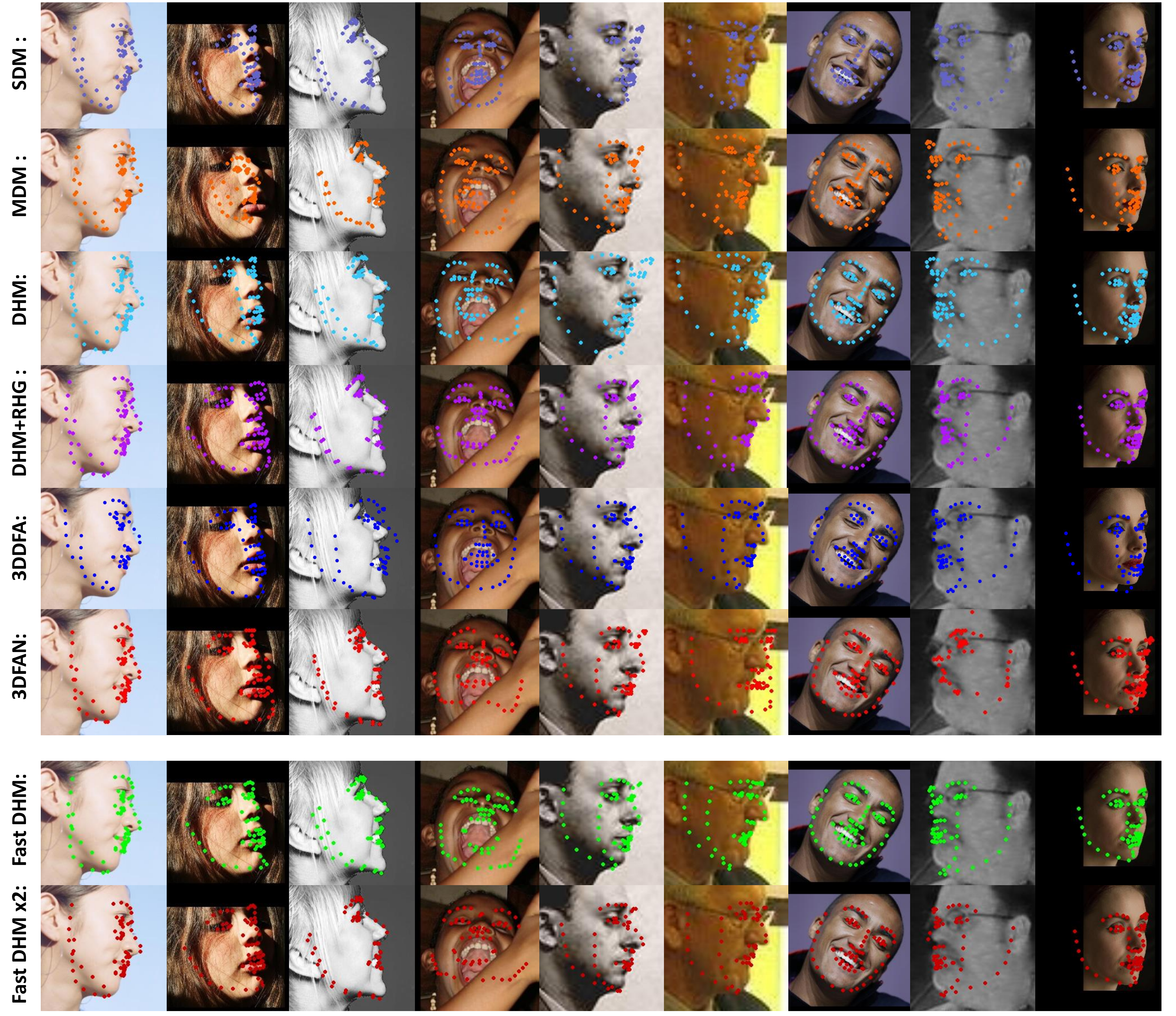}
\end{center}
   \caption{Illustrations of alignment results by different methods. The bottom two rows of the figure show the results of Fast DHM$\times 2$ and Fast DHM.}

\label{fig:visualresults}
\end{figure*}

\subsection{Baselines and Parameter Settings}

 We conduct comprehensive evaluations with state-of-the-art methods. In this paper, a comparison is made with deep state-of-the-art methods PCD-CNN \cite{kumar2018disentangling}, 3D-FAN \cite{bulat2017far}, Hyperface \cite{hyperface}, 3DSTN \cite{3DSTN}, 3DDFA  \cite{zhu2016face}, and MDM \cite{trigeorgis2016mnemonic}. Among these baselines, 
the results of 3DSTN and PCD\text{-}CNN are cited from their original papers. We also compare the results and speed on CPU with some traditional methods running on CPU, including SDM \cite{xiong2013supervised}, ERT \cite{kazemi2014one}, and ESR \cite{esr}. The brief introduction of each method is:
\begin{table*}

\setlength{\tabcolsep}{7pt}
\centering
\caption{Comparisons with state-of-the-art methods. We highlight the top three results in each setting. Fast DHM$\times 2$ means the channels in the fast CNN module are doubled.  }
\begin{tabular}{|c|c|c|c|c|c|c|c|c|}
 \hline
 
 \multicolumn{5}{|c|}{Normalized Mean Error on AFLW2000-3D} & \multicolumn{2}{c|}{Speed (FPS)}& Memory \\
 \hline
 Method Name&$[0^\circ 30^\circ]$&$[30^\circ 60^\circ]$&$[60^\circ 90^\circ]$&Mean&GPU &CPU &params (Bytes)\\
 \hline
  \textbf{Fast DHM$\times 2$}  &\textbf{2.43}&\textbf{3.51}&5.41 & \textbf{3.78}&\textbf{740}&53&\textbf{2.2M}\\
  \textbf{Fast DHM}&\textbf{2.57}&3.60&5.88 &4.01&\textbf{$>$900}&\textbf{81}&\textbf{0.9M}\\
  DAMDNet ({\text{2019ICCVW}}) \cite{DAMDNet} &2.91&3.83&\textbf{4.95} & 3.89&$50$&$12.5$&11.3M\\
  Improved 3DDFA ({\text{2019TPAMI}}) \cite{zhu2019face} &2.84&3.57&\textbf{4.96} & \textbf{3.79}&$45$&$8$&12.6M\\
DHM+RHG ({\text{2018BMVC}}) \cite{Sun2018DeepE3} &\textbf{2.52}&\textbf{3.21}&5.48 & 3.85&15&$<1$&221.9M\\
DHM ({\text{2018BMVC}}) \cite{Sun2018DeepE3} &2.75&4.21&6.91  & 4.11 &27&$<1$&193.7M\\
PRNet ({\text{2018ECCV}}) \cite{PRNet} &2.75&\textbf{3.51}&\textbf{4.61}  & \textbf{3.62}& \textbf{100}&$5$&153M\\
Hyperface ({\text{2017TPAMI}}) \cite{hyperface} &3.93&4.14&6.71  & 4.26&-&-&119.7M\\
3DSTN ({\text{2017ICCV}}) \cite{3DSTN} &3.15&4.33&5.98  & 4.49&52&$<1$&-\\
 3DFAN ({\text{2017ICCV}}) \cite{bulat2017far}&2.75&3.76&5.72  &4.07&6&$<1$&183M\\
 3DDFA ({\text{2016CVPR}}) \cite{zhu2016face} &3.78&4.54&7.93  & 5.42&15&8&111M\\
3DDFA+SDM ({\text{2016CVPR}}) \cite{zhu2016face}&3.43&4.24&7.17  & 4.94&10&7&121M\\
 MDM ({\text{2016CVPR}}) \cite{trigeorgis2016mnemonic} &3.67&5.94&10.76 &6.45 &5&$<1$&307M\\
 ERT ({\text{2014CVPR}}) \cite{kazemi2014one}&5.40&7.12&16.01    &10.55 &-&\textbf{300}&95M\\
 ESR ({\text{2014IJCV}}) \cite{esr}&4.60&6.70&12.67 &7.99 &-&\textbf{83}&248M\\
  SDM ({\text{2013CVPR}}) \cite{xiong2013supervised}&3.67&4.94&9.76   &6.12 &-&80&\textbf{10M}\\
\hline
\end{tabular}

\label{table:comparison}
\end{table*}

\begin{itemize}
\item DAMDNet \cite{DAMDNet}: DAMDNet is the state-of-the-art methods for large pose face alignment proposed in 2019. It implemented dual attention into the densenet framework and achieved impressive accuracy and speed.

\item PRNet \cite{kumar2018disentangling}: PRNet is the methods for 3D face reconstruction and alignment proposed in 2018. With the shape
constraint imposed by UVMap, it improved the performance significantly.
\item Hyperface \cite{hyperface}: Hyperface is a framework for multitask, including face alignment. Its main idea is to fuse different features by concatenation operation and usefully connected layer to the decoder the features for different tasks.
\item 3DSTN \cite{3DSTN}: 3D Spatial Transformer Networks utilize both the true 3D
model of the subject in the image and the properties of the camera used to capture the image to model how the face changes from one viewpoint to another.
\item 3DFAN \cite{bulat2017far}: 3DFAN is another state-of-the-art algorithm for 3D face alignment. It uses the Hour Glass (HG) Network to do both 2D and 3D alignment. In their structure, it has four HGs, and all the bottleneck blocks in HGs were replaced with a hierarchical, parallel and multi-scale block. In this way, it has achieved an impressive performance. 
\item 3DDFA \cite{zhu2016face}: 3DDFA is one of the state-of-the-art algorithms for 3D face alignment, using cascaded CNN as its main structure. It learns a set of CNN structure with the dense 3D features. The performance is further improved recently \cite{zhu2019face}.
\item MDM \cite{trigeorgis2016mnemonic}: MDM is the state-of-the-art algorithm for 2D alignment. The MDM regresses the patch features with the CNN structure and uses RNN to improve the cascade learning process.
\item ERT \cite{kazemi2014one}: ERT is a popular algorithm using regression trees using cascade to learn a set of weak regressors. It uses the pixel-wise feature and has achieved outstanding performance on 2D alignment.
\item SDM \cite{xiong2013supervised}: SDM is a method calculating the descent in feature space to minimize the error. It also uses cascade to learn a set of linear regressors using the features and the offset of the landmarks. 
\end{itemize}

\begin{figure}[t]
\begin{center}
\includegraphics[width=1\linewidth]{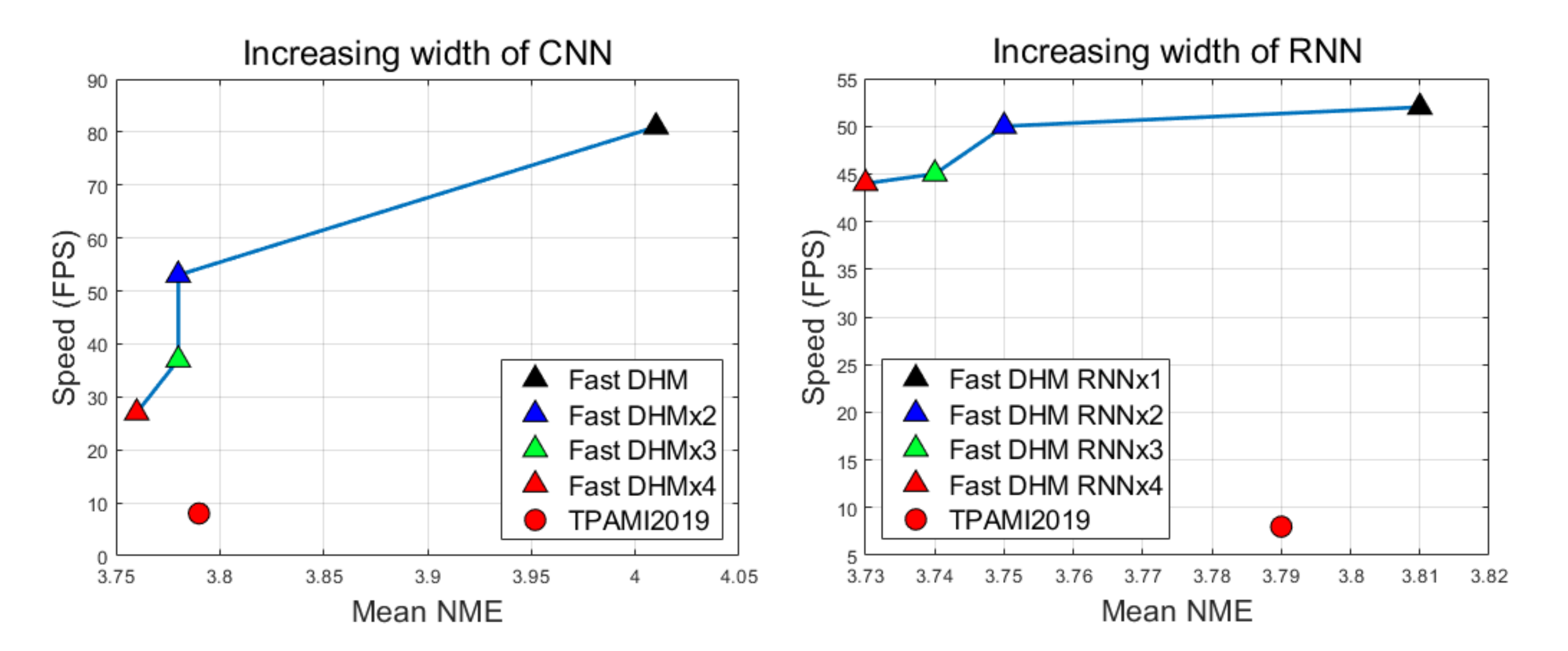}
\end{center}
\caption{Relationships of speed and mean NME among different sizes of fast CNN module (left) and fast RNN module (right) in our fast DHM framework. $\times n$ means the channels of fast CNN module is extended by $n$. We also show the result of TPAMI 2019 \cite{zhu2019face} for clear comparison of the baseline. Note that the experiments of fast RNN module is implemented based on Fast DHM CNN $\times 2$. }
\label{fig:chartspeed}
\vspace*{-3pt}
\end{figure}

To learn the weights of the network, we use Adam stochastic optimization \cite{kingma2014adam} with default hyperparameters. The initial learning rate is 0.001 for 300W-LP with exponential decay of 0.95 every 2000 iterations, and an initial learning rate of 0.0001 is employed in our training process. The batch size is set to 50. The Recurrent HourGlass (HG) network starts with a $7\times7$ convolutional layer with stride 2. A residual module and a round of max pooling are added after it to bring the resolution down from 256 to 64. We use 3 stacked HG modules to extract features and an LSTM \cite{lstm} as our recurrent module. The initial learning rate is 0.001, and we set weight decay at epoch $5$, $15$, $30$. The total number of epochs is $40$. We use RMSprop \cite{rmsprop} as our optimizer. The training batch is $32$, and validation batch is $16$.

Fast DHM framework is trained with $0.005$ as its initial learning rate. The optimization method is also Adam stochastic optimization. The training batch size is set to $50$. Training epoch is set to $120$. After $60$ epochs, the learning rate is reset to $0.00001$.

\begin{figure}[t]
\begin{center}
\includegraphics[width=1\linewidth]{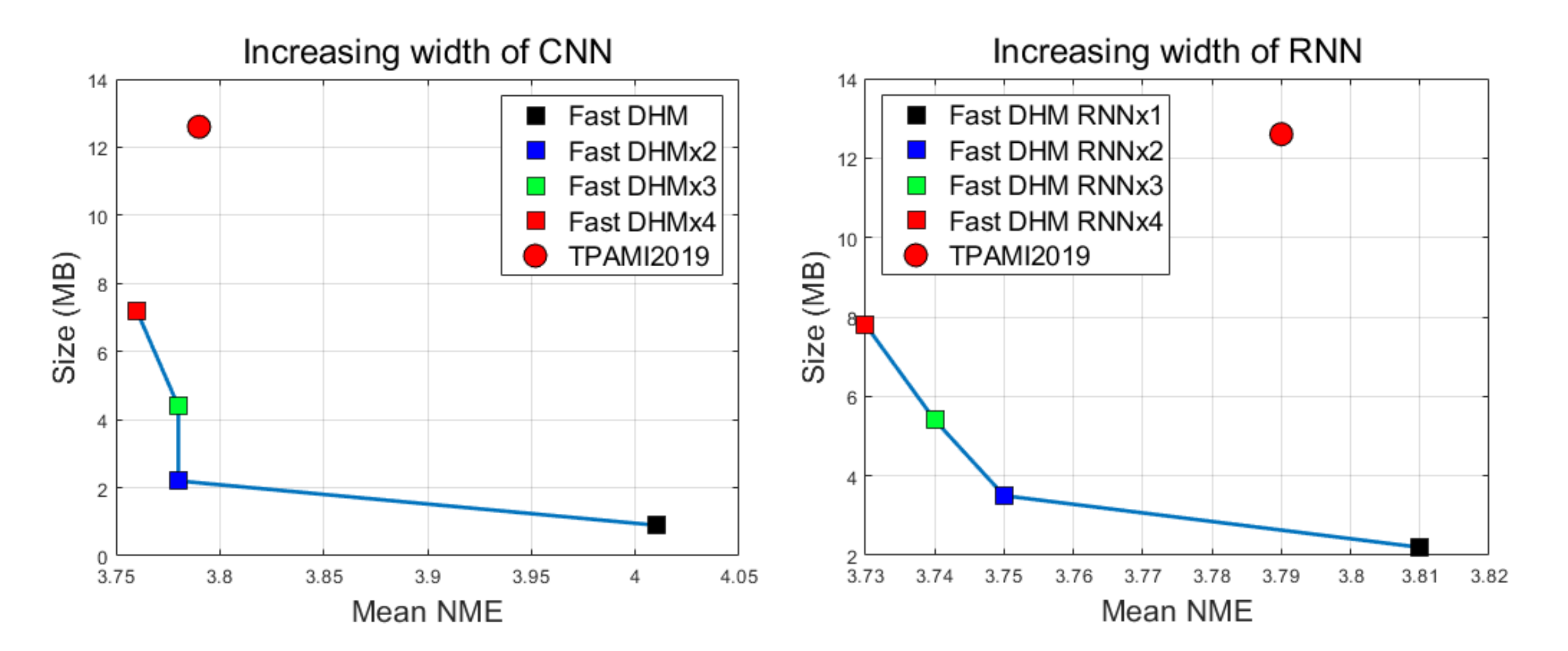}
\end{center}
\caption{Relationships of storage size and mean NME among different sizes of fast CNN module (left) and fast RNN module (right) in our fast DHM framework. $\times n$ means the channels of fast CNN module is extended by $n$. We also show the result of TPAMI 2019 \cite{zhu2019face} for clear comparison of the baseline. Note that the experiments of fast RNN module is implemented based on Fast DHM CNN $\times 2$.}
\label{fig:chartsize}
\vspace*{-3pt}
\end{figure}

\subsection{Performance Analysis of Our Model}
In this section, we will demonstrate the advantage and necessity of two modules: (1) 3D diffusion heat maps; (2) RNN for deep evolution.

First, we evaluate the importance of evolutionary 3D maps. Instead of using RNN, we use a plain 3D-CNN structure. That is being said, we use the same 3D heat maps but replace the RNN by a conventional CNN structure. The rest parts remain the same. Note we use 300W-LP dataset for training and AFLW2000-3D dataset for evaluation. Results of this setting (3D-CNN) can be found in Figure \ref{fig:2DRNN&3DCNN&all}(a).

Second, we keep the RNN structure and test the importance of 3D heat maps in our framework. We replace the 3D module by initial 2D landmarks. Accordingly, we change the output of the framework from a $234\times 1$ vector to a $204\times 1$ vector, which is the increment of locations of predicted landmarks. The rest of the framework remains the same. Note we also use the same training and testing datasets. The result of this setting (2D-RNN) can be found in Figure \ref{fig:2DRNN&3DCNN&all}(a). It can be found that the combination of both modules performs best.

\subsection{Comparisons with Existing Methods}

In this section, we conduct comprehensive evaluations with state-of-the-art methods. Especially, all methods are trained on the 300W-LP dataset including both ours and others. All of the input faces are cropped by the bounding box calculated from landmarks. The methods with released codes are trained on the 300W-LP and the comparison CED curves can be found in Figure \ref{fig:2DRNN&3DCNN&all} (b) to (h). The quantitative results can be found in Table \ref{table:comparison}. 

\textbf{Accuracy:} In the Table \ref{table:comparison}, we highlight the top three methods in each columns. Our proposed fast deep evolutionary network has the second lowest mean NME among all the methods. Although our NME in $[60^\circ 90^\circ]$ is $0.8$ higher than PRNet and is not in the top three, our performance ranks top three in $[0^\circ 30^\circ]$, $[30^\circ 60^\circ]$ and mean error evaluation, which means our fast deep evolutionary framework with DHM is competitive with the state-of-the-art methods on the accuracy. From the experiments, we can prove that our approach is very robust and reliable on different datasets. The visualization of eight methods are shown in Figure \ref{fig:visualresults}.

\textbf{Time Complexity:} Since face alignment algorithms are often used on mobile devices, real-time implementation of the algorithms without GPU support is important. Compared with those deep learning methods \cite{DAMDNet,zhu2019face,hyperface,3DSTN,zhu2016face,bulat2017far,hinton2015distilling,trigeorgis2016mnemonic}, our fast deep evolutionary framework has much better speed on both one core CPU and GPU. Our speed is $\times6$ of DAMDNet on CPU and $\times14$ on GPU. Besides, our Fast DHM$\times2$ is $\times10$ faster than PRNet on CPU and $\times7$ faster on GPU. The results are shown in Table \ref{table:comparison}. There are two reasons for the impressive acceleration. First, the lightweight depthwise convolution is used in the CNN structure and the dimension of each layer is small. Second, the recurrent block is replaced by only one pointwise layer and two depthwise layers.  In the table, we notice that the SDM \cite{xiong2013supervised}, ERT \cite{kazemi2014one} and ESR \cite{esr} have very impressive speed on CPU. The reason is all of these methods use hand-craft features which are computationally efficient. However, the accuracy based on these features is inferior and fail to provide great representation. We also explore the relationship between the speed and mean NME among different sizes of fast CNN module and RNN module relatively. The results are shown in Figure \ref{fig:chartspeed}. From the figure, it is clear to see that there is a saturate point when the mean NME is around $3.8$, which can be reached by just double the channels of fast CNN module. After the saturate point, the loss of the speed is far beyond of the benefit from the improvement of the accuracy.  However, the trade-off between the speed and the increment of the accuracy brought by RNN module seems acceptable.

\textbf{Space Complexity:} For the applications on mobile devices, the face alignment model may use limited memory for functionality. Thus, it is important to measure the memory consumption of each model. From Table \ref{table:comparison}, it can be observed that the DHM with the fast evolutionary framework is $\times 12.5$ smaller than the smallest model in baseline deep learning methods. Besides, it is $\times 10$ smaller than the smallest model in all baselines. Note that, since the DHM is pre-defined before the testing stage, we do not take the size of BFM model into consideration. We also show the relationship between the size and mean NME among different sizes of fast CNN module in Figure \ref{fig:chartsize}. From the figure, we can observe that our Fast DHM reaches the saturate point at the Fast DHM $\times 2$, which means the channels of the fast CNN module are doubled. For RNN module, the reduction of mean NME is only $0.03$ while the storage size increases almost $50 \%$ every time the RNN module is increased, which is not cost-effective.
Thus, the Fast DHM $\times 2$ is utilized to achieve the best performance on mean NME, speed, and model size.

\begin{figure}[t]
\begin{center}
\includegraphics[width=1\linewidth]{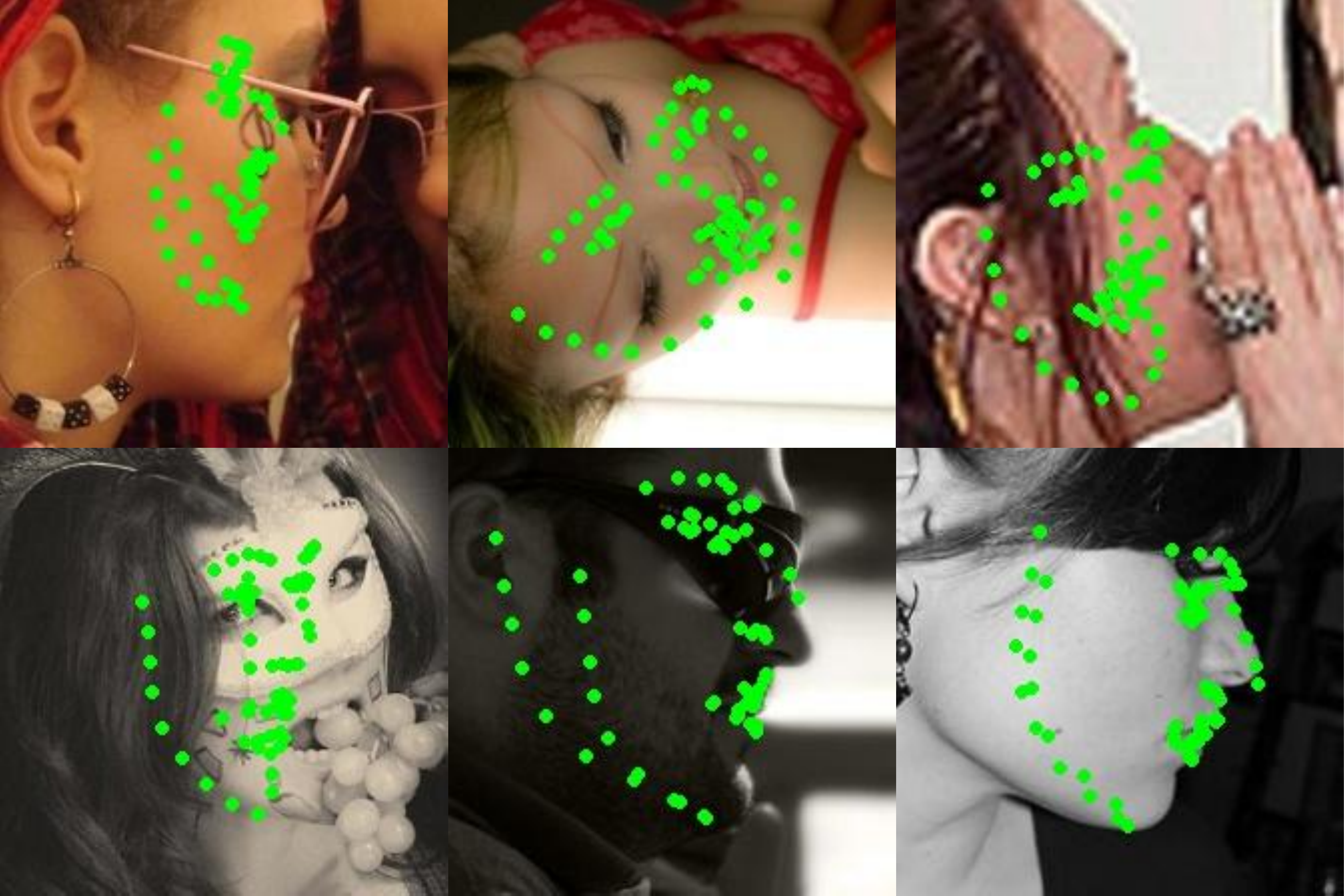}
\end{center}
\caption{Illustration of some failure cases by our models. The errors mainly occur when the large-pose faces are occluded.}
\label{fig:failcase}
\vspace*{-3pt}
\end{figure}

We also illustrate some failure cases of our model in Figure \ref{fig:failcase} which indicate that our model may be fragile given large-angle rolled faces with heavy occlusions. The primary reason is the lack of faces in similar cases in the training dataset, which is a common issue for all other methods. Possible solutions include adding corresponding training data to approach the special case, employing multiple initializations to avoid bad local minimal, and utilizing advanced loss to fit the data.

\section{Conclusions}
In this paper, we focused on improving the face alignment algorithms with the sparse 3D landmarks to approach the challenge of large poses. We presented a deep evolutionary framework to progressively update the 3D heat maps to generated target face landmarks. First, we proposed to use as a robust representation. Second, we demonstrated that an RNN based evolutionary learning paradigm was able to model the dynamics of least square problems and optimize the facial landmarks. Third, we proposed a fast framework for deep evolutionary learning and a lightweight recurrent block. The results show the efficiency of our framework on large pose face alignment task, and also show the possibility of further improvement with complicated structures in our evolutionary DHM strategy. At last, we believe the fast recurrent block can be further implemented on other temporal tasks.

\bibliographystyle{IEEEtran}
\bibliography{egbib}


%





\ifCLASSOPTIONcaptionsoff
  \newpage
\fi


\begin{IEEEbiography}[{\includegraphics[width=1in,height=1.25in,clip,keepaspectratio]{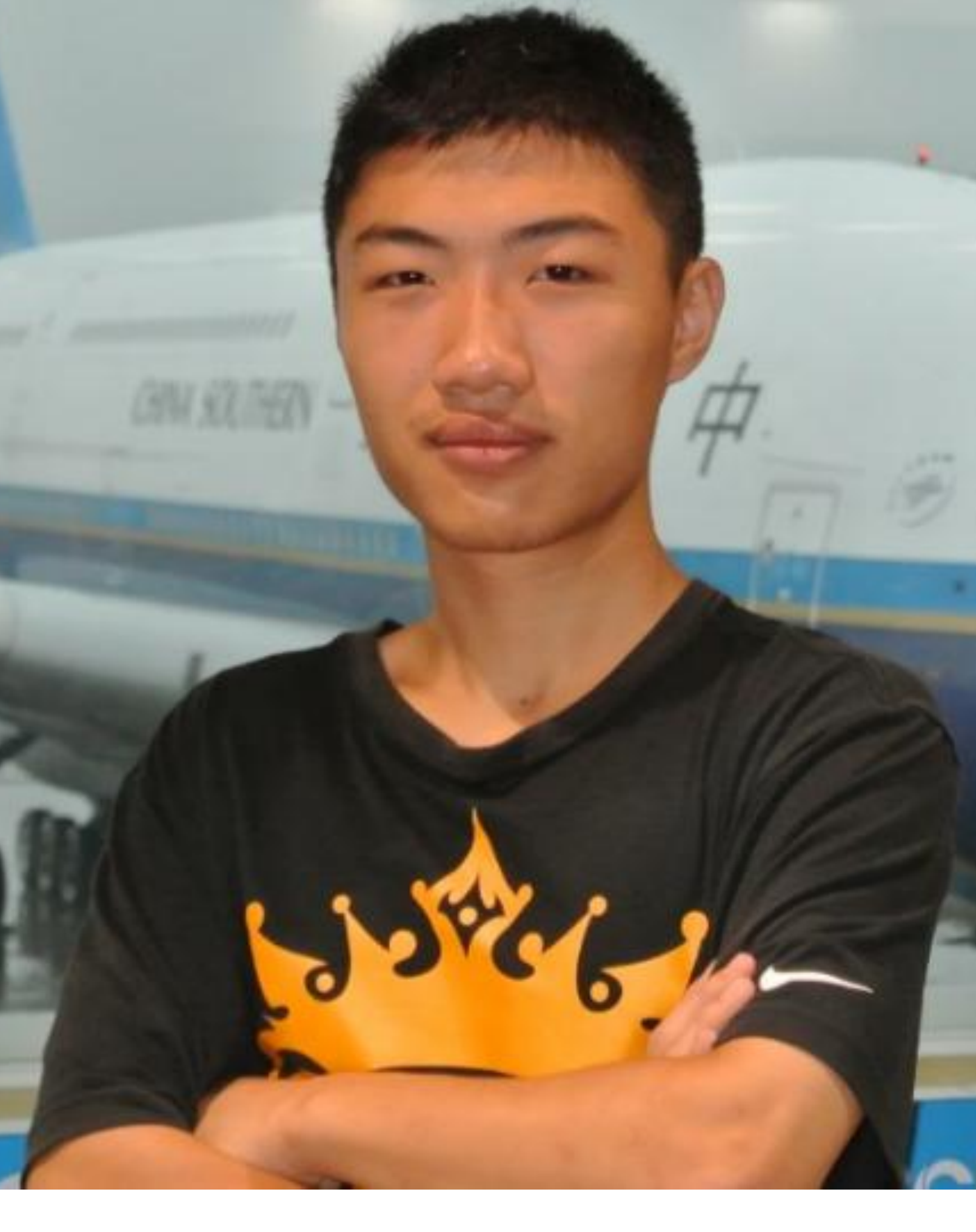}}]{Bin Sun} received the B.Eng. degree in photoelectric information engineering from Beijing Institute of Technology (BIT), China, in 2015. He is pursuing the Ph.D. degree with the Department of Electrical and Computer Engineering, Northeastern University, Boston, MA, USA. His research interests include Face Synthesis and light weight network design. He has served as PC member for AAAI, and published paper in BMVC 2018.

\end{IEEEbiography}

\begin{IEEEbiography}[{\includegraphics[width=1in,height=1.25in,clip,keepaspectratio]{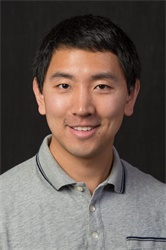}}]{Ming Shao} (S11-M16) received the B.E. degree in computer science, the B.S. degree in applied mathematics, and the M.E. degree in computer science from Beihang University, Beijing, China, in 2006, 2007, and 2010, respectively. He received the Ph.D. degree in computer engineering from Northeastern University, Boston MA, 2016. He is a tenure-track Assistant Professor affiliated with College of Engineering at the University of Massachusetts Dartmouth since 2016 Fall. His current research interests include sparse modeling, low-rank matrix analysis, deep learning, and applied machine learning on social media analytics. He was the recipient of the Presidential Fellowship of State University of New York at Buffalo from 2010 to 2012, and the best paper award winner/candidate of IEEE ICDM 2011 Workshop on Large Scale Visual Analytics, and ICME 2014. He has served as the reviewers for many IEEE Transactions journals including TPAMI, TKDE, TNNLS, TIP, and TMM. He has also served on the program committee for the conferences including AAAI, IJCAI, CVPR, ICCV, ICLR, etc. He is the Associate Editor of SPIE Journal of Electronic Imaging, and IEEE Computational Intelligence Magazine. He is a member of IEEE.
\end{IEEEbiography}

\begin{IEEEbiography}[{\includegraphics[width=1in,height=1.25in,clip,keepaspectratio]{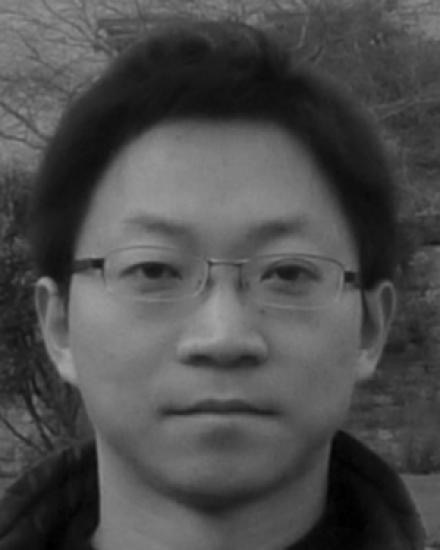}}]{Siyu Xia}
received his BE and MS degrees in automation engineering from Nanjing University of Aeronautics and Astronautics, Nanjing, China, in 2000 and 2003, respectively, and the PhD degree in pattern recognition and intelligence system from Southeast University, Nanjing, China, in 2006. He is currently working as an associate professor in the School of Automation at Southeast University, Nanjing, China. His research interests include object detection, applied machine learning, social media analysis, and intelligent vision systems. He was the recipient of the Science Research Famous Achievement Award in Higher Institution of China in 2015. He has served as the reviewer of many journals including TIP, T-SMC-B, T-IFS, T-MM, IJPRAI, and Neurocomputing. He received Outstanding Reviewer Award for Journal of Neurocomputing in 2016. He has also served on the PC/SPC for the conferences including AAAI, ACM-MM, ICME, and ICMLA. He is a member of IEEE and ACM.
\end{IEEEbiography}

\begin{IEEEbiography}[{\includegraphics[width=1in,height=1.25in,clip,keepaspectratio]{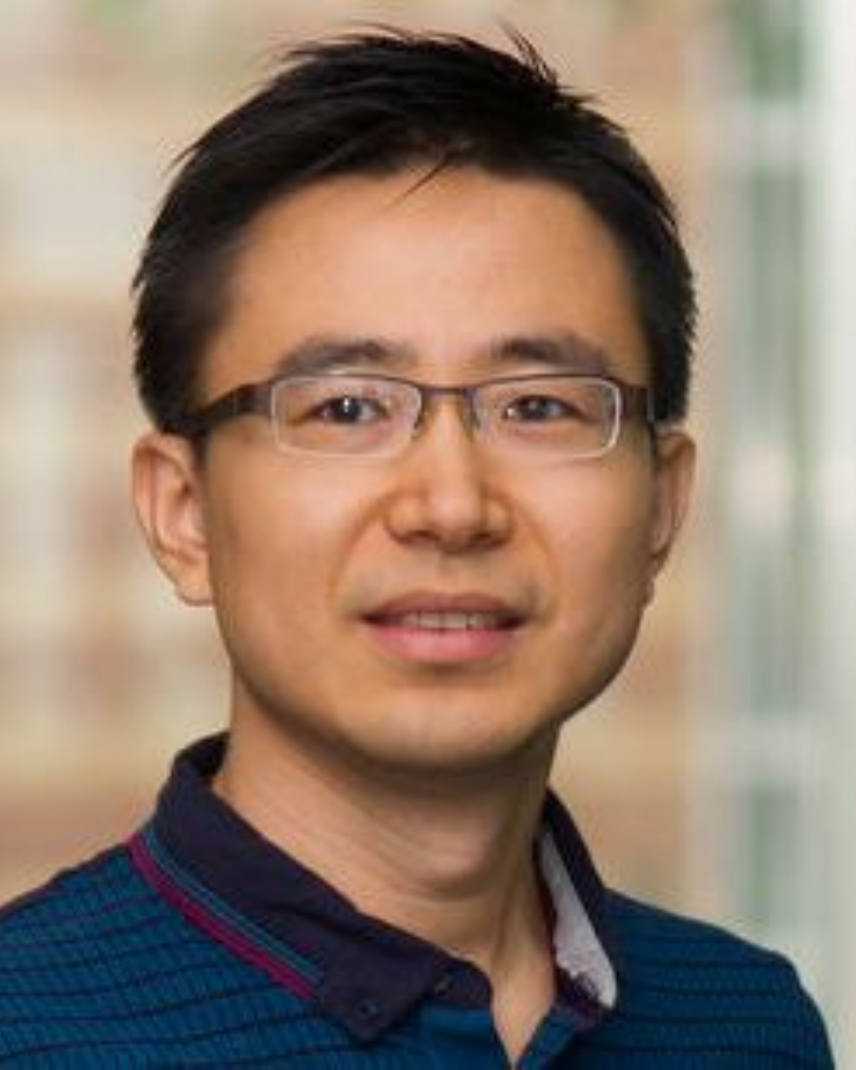}}]{Yun Fu} (S'07-M'08-SM'11-F'19) received the B.Eng. degree in information engineering and the M.Eng. degree in pattern recognition and intelligence systems from Xi’an Jiaotong University, China, respectively, and the M.S. degree in statistics and the Ph.D. degree in electrical and computer engineering from the University of Illinois at Urbana-Champaign, respectively. He is an interdisciplinary faculty member affiliated with College of Engineering and the College of Computer and Information Science at Northeastern University since 2012. His research interests are Machine Learning, Computational Intelligence, Big Data Mining, Computer Vision, Pattern Recognition, and Cyber-Physical Systems. He has extensive publications in leading journals, books/book chapters and international conferences/workshops. He serves as associate editor, chairs, PC member and reviewer of many top journals and international conferences/workshops. He received seven Prestigious Young Investigator Awards from NAE, ONR, ARO, IEEE, INNS, UIUC, Grainger Foundation; nine Best Paper Awards from IEEE, IAPR, SPIE, SIAM; many major Industrial Research Awards from Google, Samsung, and Adobe, etc. He is currently an Associate Editor of the IEEE Transactions on Neural Networks and Leaning Systems (TNNLS). He is fellow of IEEE, IAPR, OSA and SPIE, a Lifetime Distinguished Member of ACM, Lifetime Member of AAAI and Institute of Mathematical Statistics, member of ACM Future of Computing Academy, Global Young Academy, AAAS, INNS and Beckman Graduate Fellow during 2007-2008.
\end{IEEEbiography}

\end{document}